\pdfoutput=1
\documentclass[11pt]{article}

\usepackage{ACL2023}

\usepackage{times}
\usepackage{latexsym}

\usepackage[T1]{fontenc}
\usepackage[utf8]{inputenc}

\usepackage{microtype}

\usepackage{inconsolata}

\usepackage{enumitem}
\usepackage{graphicx}
\usepackage{booktabs}
\usepackage{tabularx}
\usepackage{multirow}

\title{Identity Construction in a Misogynist Incels Forum}

\author{Michael Miller Yoder\textmd{\textsuperscript{1}}, Chloe Perry\textmd{\textsuperscript{2}}, David West Brown\textmd{\textsuperscript{3}}, \\ \textbf{Kathleen M. Carley}\textsuperscript{1}, \textbf{Meredith L. Pruden}\textsuperscript{4}\\
    \textsuperscript{1}Software and Societal Systems Dept., Carnegie Mellon University, Pittsburgh, PA, USA \\
    \textsuperscript{2}Dept. of American Culture, University of Michigan, Ann Arbor, MI, USA \\
    \textsuperscript{3}Dept. of English, Carnegie Mellon University, Pittsburgh, PA, USA \\
    \textsuperscript{4}School of Communication and Media, Kennesaw State University, Kennesaw, GA, USA \\
    \texttt{yoder@cs.cmu.edu}, \texttt{chloeper@umich.edu}, \texttt{dwb2@andrew.cmu.edu}, \\ 
    \texttt{carley@cs.cmu.edu},
    \texttt{mpruden@kennesaw.edu} \\
}

\begin{document}
\maketitle
\begin{abstract}
Online communities of involuntary celibates (incels) are a prominent source of misogynist hate speech.
In this paper, we use quantitative text and network analysis approaches to examine how identity groups are discussed on incels.is, the largest black-pilled incels forum.
We find that this community produces a wide range of novel identity terms and, while terms for women are most common, mentions of other minoritized identities are increasing.
An analysis of the associations made with identity groups suggests an essentialist ideology where physical appearance, as well as gender and racial hierarchies, determine human value.
We discuss implications for research into automated misogynist hate speech detection.

\end{abstract}

\section{Introduction}
\textbf{\textit{Warning: this paper contains content that is disturbing, offensive, and/or hateful.}}

Online communities of those calling themselves ``involuntary celibates'' (incels) are known for online misogynist hate speech and offline violence targeting women, including incidents of mass violence in Isla Vista, California, in 2014 and Toronto, Canada, in 2018, among others.
Though some work in natural language processing (NLP) has focused on features of misogynist language in general \citep{anzovino_automatic_2018,samghabadi_aggression_2020,guest_expert_2021}, online incel communities are known for significant lexical innovation \citep{farrell_use_2020,gothard_incel_2021}.
Training with data from incel forums would enable misogynist hate speech classifiers to identify the neologisms and novel ideological features of this dangerous form of online misogyny~\citep{jaki_online_2019}.

In this paper, we provide hate speech researchers with a quantitative overview of trends and particularities of language in one of the largest misogynist incel communities, incels.is, which launched in 2017 following the r/incels ban from Reddit.
We focus this analysis on mentions of identities, which are key to automatically identifying hate speech~\citep{uyheng_identity-based_2021} and a window into the ideologies of social movements~\citep{benford_framing_2000}.
We ground this analysis in theoretical approaches that focus on how identities are \textit{constructed} in interaction~\citep{bucholtz2005identity,burr_social_2017} and investigate the following research questions:

\begin{quote}
    \begin{enumerate}[label=\textbf{RQ\arabic*.}]
        \item How frequently are different identities, including novel terms for identities, mentioned in incels.is discourse? 
        \item How do identity mentions in incels.is discourse change over time?
        \item How are identity mentions used differently by central incels.is users?
        \item What textual associations are made with identity groups on incels.is?
    \end{enumerate}
\end{quote}

To address these research questions, we first measure the distribution of identity term mentions using a large generic list of identity terms combined with community-specific identity terms surfaced from a word embedding-based approach.
We confirm the most frequent identity mentions in this data are for women, with almost one-fourth of these being derogatory community-specific neologisms, such as ``femoids.''
Mentions of gender are much higher than in a comparative white supremacist dataset, a similar commonly-used source of unlabeled hate speech~\citep{simons_bootstrapped_2020,alatawi_detecting_2021}.
We find increasing mentions of other minoritized identities, such as Black, LGBTQ+ and Jewish people on incels.is, suggesting a consolidation with broader far-right discourses.
Users who are central to the network proportionally mention more of these other marginalized identities. From a quantitative analysis of the immediate contexts in which identity term mentions appear, a pervasive hatred of women is clear, as well as reinforcement of stereotypes about other marginalized groups. The incel identity itself is often discussed with themes of victimhood and boundary-keeping for ``true'' versus ``fake'' incels.

Throughout our analysis, we find evidence of an essentialist, black-pilled ideological framework where physical appearance determines the value of individuals and groups.
While rigid racial and gender hierarchies are not new (e.g., eugenics) and are often circulated in far-right discourse~\citep{miller-idriss_hate_2022}, this incel community attaches many novel measurements to appearance related to these hierarchies, re-entrenching and extending them.  
We argue that to detect such a particular form of extremism, hate speech researchers must heed both the jargon and deeper ideologies of this movement.

\section{Incels and Male Supremacism}
Online misogynist incel communities are situated within a set of anti-feminist groups often termed the ``manosphere.'' 
These groups include Men's Rights Activists, Pick Up Artists (PUAs), Men Going Their Own Way (MGTOW).
Such groups are often associated with a ``red pill'' ideology, a \textit{Matrix} film reference to seeing the hidden truth behind the world, in this far-right context a view that feminism has brainwashed and subordinated men~\citep{ging_alphas_2019}.
In addition, incels often refer to a ``black pill,'' the idea that they are genetically pre-determined to be incels and cannot improve their situation through work or self-improvement~\citep{pruden_maintaining_2021}.
This leaves many black-pilled incels feeling that their only options are to cope, commit suicide, or commit mass violence (expressed in the common phrase, ``cope, rope or go ER [Elliot Rodger, an incel mass shooter]).'' 
Among groups in the manosphere, incels are most associated with violent and high-profile events that demonstrate ``extreme misogyny''~\citep{ging_alphas_2019}. 

\citet{ribeiro_evolution_2021} find that incel communities are both more extreme and more popular than older, more moderate male supremacist movements, while \citet{Laviolette2019} find more extreme manosphere movements contain essentialist, deterministic ideologies of identity, which we also find in incels.is. 
We find evidence of this reductionist and biologically essentialist worldview in associations with identities in incels.is.  

Qualitative research on male supremacist extremism frequently examines Men's Rights Activists~\citep{berger_extremism_2018} and more moderate groups that are less niche and without the emergent vocabulary common in incel spaces. We find this tendency extends to the data sources in automated hate speech research and argue for the importance of attending to the particularly dangerous discourse of black-pilled incels such as those on incels.is.

Quantitative and computational studies of the manosphere often focus on the unique misogynist language use of these communities.
\citet{gothard_incel_2021} and \citet{jaki_online_2019} surface incel jargon by comparing word frequencies in incel Reddit posts with subreddits and Wikipedia articles outside of the incel movement, while \citet{farrell_use_2020} find frequent incel terms not present in English dictionaries and expand their lexicon with a word embedding space. Such word frequency analysis, as well as hand-crafted lexicons, are often used to measure and study misogyny in the manosphere~\cite{heritage_mantrapcorpus_2019,farrell_exploring_2019,jaki_online_2019}.
\citet{pruden_maintaining_2021} and \citet{perry_cognitive_2021} use topic modeling to characterize narratives and map out user trajectories on incels.is and r/theRedPill, respectively.
\citet{jaki_online_2019} use word frequency analysis to study identity construction on a similar forum, incels.me, though their 6-month dataset only enables limited time-series analysis.
In contrast, our work focuses on the use and contexts of generic and community-specific identity terms beyond a sole focus on misogyny, as well as how identity term use changes over time.

\subsection{Automated misogyny detection}
In early work on automated misogyny detection, \citet{hewitt_problem_2016} and \citet{WaseemHovy2016} developed small Twitter datasets annotated for sexism.
\citet{anzovino_automatic_2018} proposed a keyword-based annotated dataset and taxonomy for misogyny detection on Twitter, with later shared NLP tasks~\citep{fersini_ibereval_2018,fersini_overview_2018,basile_semeval-2019_2019,bhattacharya_developing_2020}.
Data for these tasks came from Twitter posts and YouTube comments based on keywords, profile information, and YouTube video topics.
Such data sources may capture mainstream misogyny but miss the unique linguistic characteristics of the incel movement.

Other annotated hate speech datasets have included data from manosphere subreddits, such as r/MensRights, r/MGTOW, r/incels, and r/TheRedPill, along with many other sources of online hate speech~\citep{qian_benchmark_2019,Sap2020,mollas_ethos_2022}.
\citet{guest_expert_2021} propose a Reddit dataset annotated for misogyny by trained annotators, who would be more likely to understand community-specific jargon than crowdworkers with limited training. 
Though they include a variety of manosphere-related subreddits, absent from this dataset are banned black-pilled incel subreddits such as r/braincels, r/shortcels, and r/incels, the precursor of the more extreme incels.is.

\section{Data}
Our dataset contains 6,248,234 English-language public comments posted between the forum's creation in November 2017 and scraping in April 2021.\footnote{This dataset, without any private or identifying information, will be made available to vetted researchers upon publication of the main paper associated with it.}
It includes forum and thread names, as well as the date of posting, user names and the comment's full text. However, it does not contain images, which is a limitation.

\paragraph{White supremacist dataset}
We compare identity mentions on incels.is to another common source of unlabeled hate speech: white supremacist texts.
From a large, multi-domain, English-language white supremacist dataset~\citep{yoder_weakly_2023}, we select posts from online forums in a similar time frame as the incels data, 2015-2019 (the latest year available in the white supremacist dataset).
This subset includes 3,410,623 posts from Stormfront, Iron March, and 4chan /pol/ in threads with fascist and white supremacist topics or posted by users choosing white supremacist, Nazi, Confederate or fascist flags.

\begin{figure*}[tb]
    \centering
    \includegraphics{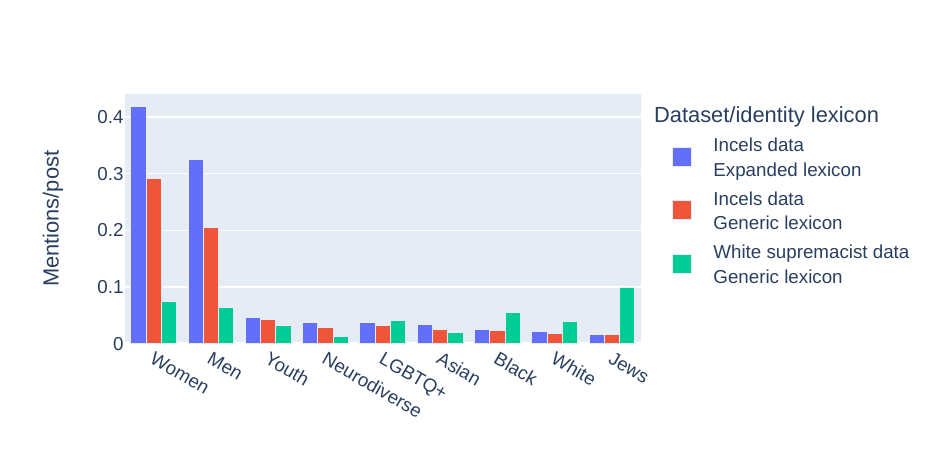}
    \vspace*{-8mm}
    \caption{Identity group mention frequencies.}
    \label{fig:identity_group_mentions}
\end{figure*}
% Plots generated in match_identities.ipynb

\section{Methods}
We take a quantitative approach to studying discursive identity construction~\citep{bucholtz2005identity,Gee2011}, borrowing a focus on in-group and out-group identity presentation from social identity theory~\citep{Tajfel1974,Seering2018}.
Specifically, we examine the use of identity terms and the immediate contexts in which they appear.
A few mentions of an identity group may not represent attitudes of participants, but associations repeatedly made over the course of a 6 million-post corpus are more likely to capture widely shared beliefs~\citep{stubbs_words_2001}.

\subsection{Measuring the use of identity terms}
We first find identity terms using a generic lexicon combined from multiple sources: the extensive list of English identity terms from the NetMapper software~\citep{joseph_social-event_2016,carley_ora_2018}, as well as identity terms frequently found in hate speech~\citep{yoder_how_2022} and terms for LGBTQ+ and neurodiverse identities found online~\citep{Yoder2020,yoder_computational_2021}.
This combined lexicon totals 19,050 unique identity terms.
Ignoring case, 7,244 were present in the incels.is dataset.

\paragraph{Grouping identity terms}
We aggregate identity terms referring to similar groups (such as \textit{LGBTQ+ people}) and then further group those identities into broader demographic categories (such as \textit{gender/sexuality)}.
To form these groupings, we adapt identity terms group labels used in hate speech research from \citet{uyheng_bots_2020} and \citet{yoder_how_2022}.\footnote{Non-proprietary portions of identity term lexicons (including groupings and categorizations) and code for analyses in this paper are available at \url{https://github.com/michaelmilleryoder/incels_identities}.}
Intersectional identity terms are counted for all groups indicated by the term, e.g., ``white women'' was counted for both ``white'' and ``women.''

\paragraph{Identity lexicon expansion} To capture the neologisms that incel communities are known for~\citep{jaki_online_2019,gothard_incel_2021}, we expand our generic identity lexicon to nearest neighbors in word embedding space, a common approach~\citep{Demszky2019,simons_bootstrapped_2020,lai_hamor_2021}.
We trained a 300-dimension word2vec model~\citep{Mikolov2013} over our data and manually examined terms appearing at least 1,000 times among the top 30 nearest neighbors by cosine distances to a) the 30 most frequent generic identity terms or b) the mean of identity term embeddings in an identity group.
This resulted in 84 new terms, the most frequent of which are in Table \ref{tab:novel_terms}.

\begin{table}[tb]
\centering
\begin{tabularx}{\columnwidth}{X}
\toprule
\multicolumn{1}{c}{\textbf{Community-specific identity terms}} \\
foids, chads, manlets, stacies, boyo, femoids, ethnics, chadlites, roasties, holes, betabux, \\
landwhales, waifus, jbs, chicks, noodlewhores, soyboy, br0, aspie, betas, thots, traps, beckies, m8, boi \\
\toprule
\multicolumn{1}{c}{\textbf{``Cel'' variants}} \\
truecels, fakecels, volcels, greycels, \\
escortcelling, gymcelling, ricecels, mentalcels, currycels, fatcels, femcels, whitecels, framecels, youngcels,
oldcels, blackcels, ethnicels, brocels, itcels, incelistan, nearcels, tallcels, shortcels, \\
locationcels, bluecels \\
\bottomrule
\end{tabularx}
\caption{Most frequent 25 novel identity  and ``cel'' terms found in the incels.is dataset.
% Plurals are combined with singular mentions and ``-ing'' terms combined with root terms.
Plural and singular mentions are combined, as are ``-ing'' terms with their roots.
}
    \label{tab:novel_terms}
\end{table}
% Generated in cel_variants.ipynb and match_identities.ipynb

% \begin{table}[tb]
% \centering
% \begin{tabularx}{\columnwidth}{X|X}
% \toprule
% \textbf{Identity terms} & \textbf{``Cel'' variants} \\
% \hline
% foids, chads, manlets, stacies, boyo, femoids, ethnics, chadlites, roasties, holes, betabux, landwhales, waifu, jb, chicks, noodlewhores, soyboy, br0, aspie, betas, thots & truecels, fakecels, volcels, greycels, escortcelling, gymcelling, ricecels, mentalcels, currycels, fatcels, femcels, whitecels, framecels, youngcels, oldcels, blackcels, ethnicels, brocels, itcels, incelistan, nearcels \\
% % foids & truecels \\ 
% % chads & fakecels \\
% % manlets & volcels \\
% % stacies & greycels \\
% % boyo & escortcelling \\
% % femoids & gymcelling \\
% % ethnics & ricecels \\
% % chadlites & mentalcels \\
% % roasties & currycels \\
% % holes & fatcels \\
% % betabux & femcels \\
% % landwhales & whitecels \\
% % waifu & framecels \\
% % jb & youngcels \\
% % chicks & oldcels \\
% % noodlewhores & blackcels \\
% % soyboy & ethnicels \\
% % br0 & brocels \\
% % aspie & itcels \\
% % betas & incelistan \\
% % thots & nearcels \\
% % \bottomrule
% \end{tabularx}
% \caption{Most frequent 20 ``cel'' variants in incels.is dataset.
% Plurals are combined with singular mentions and ``-ing'' terms combined with root terms. Counts range from 57,619 for ``truecels'' to 1604 for ``nearcels.''}
%     \label{tab:novel_terms}
% \end{table}
% % Generated in cel_variants.ipynb and match_identities.ipynb

\paragraph{Varieties of ``incels''} 
It is common in incel discourse to refer to different types of incels with terms including a ``cel'' suffix~\citep{gothard_incel_2021}.
For example, ``tallcels'' refers to tall incels and the racist terms ``currycels'' and ``ricecels'' refer to South Asian and East Asian incels, respectively.
Excluding usernames, over 1500 unique words used in our incels.is dataset contained the string ``cel,'' many of which referred to varieties of incels.
We examined the 100 most frequent words containing ``cel'' and grouped words that referred to incel variants, except those referring to ``fake'' incels, within the incels identity group for further analysis.

\paragraph{Central forum users}
We also analyze how forum leaders (prototypical incels) use identity terms. To find such leaders based on network structure, we construct a undirected graph where nodes are users and edges are weighted by the number of shared threads (out of 154,049 threads) between them.
This graph contains 6819 users and 3,889,054 links.
We operationalize central users as the top 5\% ranked by eigenvector centrality, which measures if users share threads with other highly-connected users.
These central users had a roughly similar number of posts as the rest of the users combined.

\subsection{Associations with identity terms}
Beyond the occurrence of identity term mentions, we analyze associations made with identities in their immediate contexts.
Specifically, we extract \textit{actions} taken by or to these groups, as well as \textit{attributes} associated with them, a simple approach to analyzing the presentation of entities in discourse~\citep{Bamman2013,Bamman2014,yoder_computational_2021}.
For actions, we extract verbs where an identity term is the subject or object from a dependency parse.
Attributes are adjectives and appositives whose head word is an identity term.

We surface the actions and attributes most distinctively associated with each identity group with PMI\textsuperscript{3}~\citep{daille_approche_1994,role_handling_2011}, a variant of pointwise mutual information that lowers the ranking of low-frequency terms.

\section{Results}

\begin{figure*}[tb]
\resizebox{\textwidth}{!}
    {\includegraphics{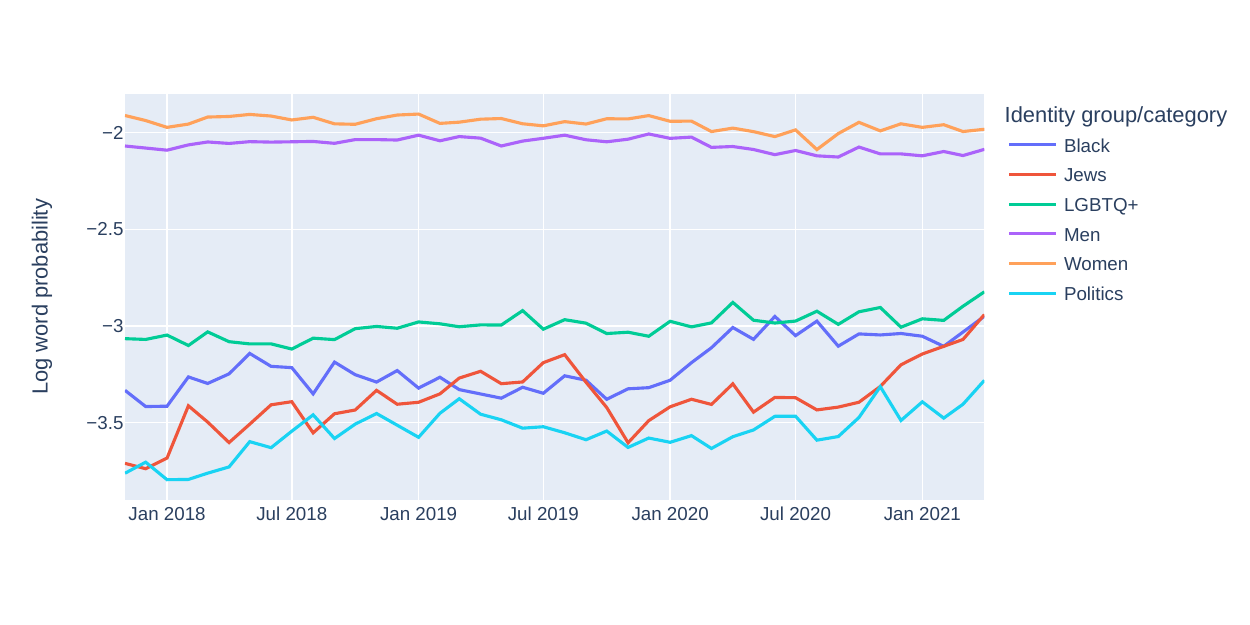}}
    \vspace*{-15mm}
    \caption{Selected identity group and category mentions over time in the incels.is dataset. Mentions of other identity groups remain steady.}
    \label{fig:identity_group_mentions_overtime}
\end{figure*}

\subsection{Distribution of identity mentions (RQ1)}
Prevalence of the most popular identity group mentions in our incels.is dataset is seen in Figure \ref{fig:identity_group_mentions}.
Expanding the generic identity list with context-specific identity terms dramatically increases the detection of mentions of all identity groups, especially women, men, and neurodiverse people\footnote{Many incels self-identify on the autism spectrum.}.
Adding these context-specific identity terms increases the total number of mentions identified from 6.46 million to 8.91 million--a jump of 37.8\%--demonstrating how common identity term innovations are in this community.

Also visible in Figure \ref{fig:identity_group_mentions} is a comparison of identity group mention frequency with another common source of hate speech, white supremacist data.
Mentions of women and men are much more frequent in the incel data than the white supremacist data, surpassing 0.4 mentions/post for women. Mentions of racial identities and Jewish people are more commonly found in the white supremacist data.
This confirms that discourse from incel communities can be a useful source of misogynist text, especially after recognizing the lexical innovations referring to women and others.

\subsection{Identity mentions over time (RQ2)}
Figure \ref{fig:identity_group_mentions_overtime} displays the prevalence of identity group mentions in this forum over time, binned every month during the dataset range and identified with the expanded lexicon.
To control for any systematic changes in post word count over time, we present log word probability (the logarithm of identity group mention counts normalized by total word count).

Though mentions of women and men are most frequent across the data range, they stay steady or slightly decrease over time.
There is a steady rise, however, in mentions of LGBTQ+ identities.
Except for a decrease in the latter half of 2019, mentions of Jewish people also steadily rise.
There is a significant rise in mentions of Black people in 2020, reaching a peak in June (during the anti-racist uprising against police brutality) and remaining at an increased rate through 2020 and 2021.
Mentions of political identities also rise.

\subsection{Central users' use of identity terms (RQ3)}
\begin{figure}[tb]
% \resizebox{\columnwidth}{!}{
    \includegraphics[trim={20mm 0 3mm 11mm},clip,width=\columnwidth]{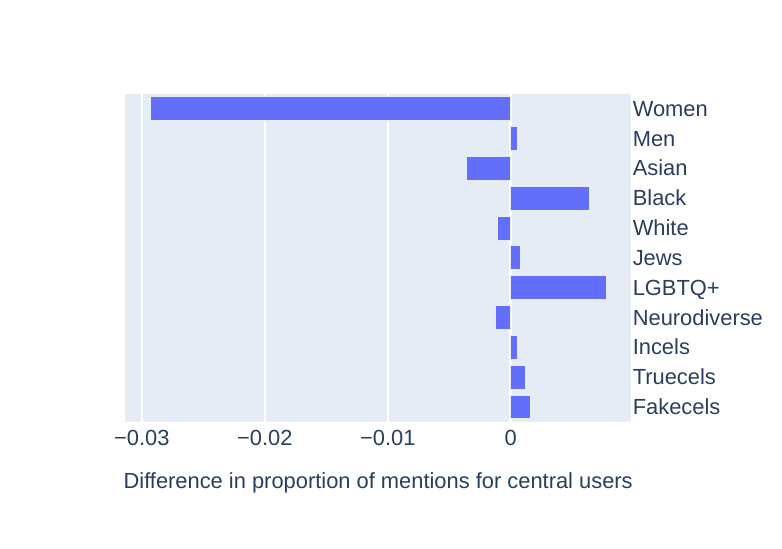}
    % }
    \vspace*{-8mm}
    \caption{Absolute difference between the proportion of identity mentions used by top 5\% central users in the shared thread network for each identity group and the proportion of mentions used by the rest of the users.}
    \label{fig:top5percent_eigen_identity_group_mentions}
\end{figure}

Figure \ref{fig:top5percent_eigen_identity_group_mentions} shows the absolute difference in proportion of identity term mentions for the top 5\% of users ranked by eigenvector centrality, compared to the rest.
Proportionally, central users are less likely to mention identity terms for women, but are more likely to mention Black, LGBTQ+, and Jewish people.
A concern for incel authenticity is also reflected in this central group's increased use of ``truecels'' and ``fakecels'' compared to other users.

\begin{table*}[tb]
% \begin{tabular}{lll}
\begin{tabularx}{\textwidth}{llX}
\toprule
\textbf{Identity} & & \textbf{Top PMI\textsuperscript{3} terms} \\
\midrule
\multirow[c]{3}{*}{Women} & \textit{Attr} & white, old, single, fat, young, hot, fucking, ugly, cute, other, ethnic, average \\
 & \textit{Act\textsubscript{S}} & want, get, love, care, go, hate, think, like, fuck, say, give, look, find, wants \\
 & \textit{Act\textsubscript{O}} & get, fuck, hate, fucking, find, getting, having, attracted, see, want, fucked \\
\cline{1-3}
\multirow[c]{3}{*}{Men} & \textit{Attr} & ugly, white, other, good, looking, average, tall, black, nice, short, chad, young \\
 & \textit{Act\textsubscript{S}} & get, looks, go, need, look, think, want, got, mogs, fuck, going, become, gets \\
 & \textit{Act\textsubscript{O}} & over, see, know, want, fuck, hate, fucking, love, seen, get, against, laid, date \\
\cline{1-3}
\multirow[c]{3}{*}{Asian} & \textit{Attr} & south, central, >, east, average, half, ugly, other, skinned, northern, full \\
 & \textit{Act\textsubscript{S}} & look, hate, get, cope, worship, go, need, mog, tend, eat, seem, make, want, take \\
 & \textit{Act\textsubscript{O}} & learning, learn, speak, hate, seen, see, mog, killed, against, over, above, know \\
\cline{1-3}
\multirow[c]{3}{*}{Black} & \textit{Attr} & north, real, west, other, fucking, man, stupid, dumb, ugly, dark, average \\
 & \textit{Act\textsubscript{S}} & get, got, commit, slay, aspire, look, developed, gon, need, go, tend, fuck, run \\
 & \textit{Act\textsubscript{O}} & free, hate, fuck, see, against, say, around, date, prefer, fucking, kill, sand \\
\cline{1-3}
\multirow[c]{3}{*}{White} & \textit{Attr} & southern, northern, non, eastern, western, pure, white, nordic, other, average \\
 & \textit{Act\textsubscript{S}} & go, get, want, mog, look, hate, invented, need, going, tend, voted, did, age \\
 & \textit{Act\textsubscript{O}} & worship, hate, prefer, against, want, date, after, over, see, sought, towards \\
\cline{1-3}
\multirow[c]{3}{*}{Jews} & \textit{Attr} & orthodox, fucking, rich, religious, secular, international, anglo, elite, greedy \\
 & \textit{Act\textsubscript{S}} & control, did, created, want, pushing, won, own, pushed, made, win, took, push \\
 & \textit{Act\textsubscript{O}} & hate, blame, ashkenazi, against, because, blaming, gas, kill, hated, hating \\
\cline{1-3}
\multirow[c]{3}{*}{LGBTQ+} & \textit{Attr} & fucking, it, bluepilled, moral, low, normal, others, larping, stupid, ill, little \\
 & \textit{Act\textsubscript{S}} & get, exist, go, say, need, think, fuck, try, deserve, trying, look, make, want \\
 & \textit{Act\textsubscript{O}} & hate, fucking, coping, fuck, shut, ban, turning, kill, banned, larping, go \\
\cline{1-3}
\multirow[c]{3}{*}{Neurodiverse} & \textit{Attr} & social, severe, functioning, fucking, extreme, crippling, sentence, complete \\
 & \textit{Act\textsubscript{S}} & exist, makes, worse, causes, affect, get, comes, means, make, sucks, goes, go \\
 & \textit{Act\textsubscript{O}} & because, due, diagnosed, cure, fucking, having, caused, cause, causes \\
\cline{1-3}
\multirow[c]{3}{*}{Incels} & \textit{Attr} & fellow, other, blackpilled, true, real, actual, white, ugly, bluepilled, blackpill \\
 & \textit{Act\textsubscript{S}} & get, exist, means, need, ascend, go, cope, know, want, say, become, going, look \\
 & \textit{Act\textsubscript{O}} & help, against, hate, coping, create, see, creating, hates, bullying, die, ok, bullied \\
\bottomrule
% \end{tabular}
\end{tabularx}
\caption{Actions and attributes associated with identity group terms in incels.is dataset. \textit{Attr} refers to attributes, while \textit{Act\textsubscript{S}} are actions for which the identity is a subject and \textit{Act\textsubscript{O}} are actions for which the identity is an object.}
\label{tab:identity_actions_attributes}
\end{table*}
% Generated in identity_associations.ipynb 

\subsection{Associations with identities (RQ4)}
Terms commonly associated with identity groups are presented in Table \ref{tab:identity_actions_attributes}.
Across groups, we find that the most frequent attributes relate to physical features (``ugly,'' ``short,'' etc.).
The use of these descriptors suggest hierarchies based on appearance, race and gender.
This focus on physical appearance is apparent in top terms used to describe women, including ``young,'' ``fat,'' and ``hot.''
Example uses show the hierarchies of appearance that incels apply to women:
``some can't tell a beta female apart from a hot whore and so lump all types of the female sub species together.''\footnote{Quotes are paraphrased for privacy~\citep{williams_towards_2017}}
Actions for which women are subjects suggest incels' speculation about what women ``want,'' ``love,''  or ``hate.''
% In other words, top attributes tend to suggest incels view women as something of a monolithic group demarcated by those who are "young" and "hot" and whom they would like "love" or "care" for and those who are "old" and "ugly" whom they "hate." 
Common actions for which women are grammatical objects include ``fuck'' and ``hate''-- evidence of the forum's misogyny, casting women as things to be ``attracted'' or controlled.
% What women want or like--as well as what they hate--is clearly a concern in this forum and vice versa. 

Men are also discussed with an emphasis on physical appearance, as well as domination.
``Ugly,'' ``looking,'' ``average,'' and ``tall'' are all top attributions for men.
Top actions include ``get,'' ``need,'' and ``mogs,'' an incel neologism meaning ``dominate''.
An example post that reinforces gender hierarchies reads, ``men literally mog femoids across the board, yet the foids whine about it.''

Race is relevant in discussions of gendered identities. 
``White,'' for example, is a top descriptor for both women and men, as is ``black'' for men. 
Top terms suggest a negotiated association of superiority with whiteness.
White people are the grammatical objects of ``worship'' but also of ``hate.'' 
``Pure'' and ``nordic,'' common white supremacist descriptors, are distinctive attributes used for white people. 
In contrast, Asian people are cast as subjects of actions like ``worship'' and ``cope.'' 

We find a range of common stereotypes for minoritized identities, particularly conspiratorial antisemitic tropes, as evidenced by terms suggesting a global Jewish conspiracy (e.g., ``elite'' and ``control'') and derogatory associations with the Holocaust (``gas'').
% One example post reads, ``The issue is that Jews control everything. That and women's rights, which we need to take away.''
One example post reads, ``the endgame is an global Jewish Communist dictatorship,'' while another mixes antisemitic conspiracy theories with anti-feminism: 
% ``Jews control porn and media, but that porn/media wouldn't have an effect if white men beat down those feminists.''
``feminism is a subversive Jewish movement designed to ruin us.''

LGBTQ+ characterizations are negative and associated with inauthenticity (e.g., ``larping,'' or live action role playing).
For example, one posts reads, ``Lesbians don't exist. They're just bisexual foids who like women but still can't resist Chad.''

Violence is associated with Black people (``commit''), for example in one post that reads, ``I don't hate blacks because they're ugly, I hate them because no matter where they are they commit crime.''

\paragraph{In-group identity associations}
Victimhood, race and authenticity are common themes associated with identity mentions of incels themselves and ``incel'' variants on the platform.

The most frequent lexical variations containing the ``cel'' suffix are in Table \ref{tab:novel_terms}.
Top terms relate to authenticity, including ``fakecel,'' ``truecel,'' and ``volcel'' (``voluntary celibate''), a focus that has also been observed in incel subreddits~\citep{gothard_incel_2021}. 
Platform affordances highlight distinctions between frequent and non-frequent posters: ``greycels'' who have posted less than 500 times have a gray-colored username and are often deemed inauthentic.
Variants related to race (``whitecels,'' ``ethnicels'') are also frequent, suggesting the importance of race in incel self-classification~\citep{jaki_online_2019,farrell_use_2020}. 
Also visible in these ``cel'' variations are a set of categories based on the familar theme of physical appearance (e.g., ``fatcels'' and ``youngcels'').
``Femcels,'' or female incels, are frequently mentioned, usually derided as outside the inherent masculinity of inceldom.

Incels are cast as merely ``existing'' or ``coping,'' (Table \ref{tab:identity_actions_attributes}) while others ``hate'' them or are ``against'' them.
This victimhood includes common masculine tropes, such as a supposed inability to control themselves.
From one post: ``we can't control what we want, devaluation of women is a coping mechanism for not being able to elicit a biological response in them.''
Common far-right narratives of victimhood at the hands of corporations, Jewish people, and the media are also present: ``Jews and the media hate incels, and the gaming industry is full of SJWs [social justice warriors].''

Race is also important--and controversial--in associations made with incels.
``White'' is a top incels attribute on the forum, and both ``ethnic'' and ``white'' are associated with truecels (see Table \ref{tab:tf_actions_attributes} in Appendix \ref{sec:appendix} for top terms related to truecels and fakecels on the forum).
There is controversy over which races occupy what positions in an assumed hierarchy, often centering around the ``just be white'' (JBW) theory that white men have access to sexual relationships with women of all races.
Some posts support this theory, e.g., 
% ``2/10 white trucel will go home to sexy asian gf while chad ethnic cries by himself'', 
``being white is a +3 when it comes to noodles [Asian women], so a 4/10 white is better than a 6/10 ethnic.''
Others challenge this notion: ``a brown man with a chiseled face will mog a white incel everywhere.'' 
Still others echo the white supremacist Great Replacement Theory, blaming JBW as a way for incels of color to ``get whitecels out so sh**skins can take over.''

``Real'' and ``true'' are top attributes associated with talk about incels, echoing a focus on authenticity in the top ``cel'' variants.
This boundary-keeping is also visible in the words associated with fakecels (e.g., ``detected'' and ``ban'').
Authentic incels are victims of women's hatred (``if women aren't trying to kill you, you're not a true incel''), post a lot (``graycels are a joke with their tiny post counts'') are unattractive (``I'm an incel, of course she said no to my hideous face''), have no female friends (``what true incel has a female friend? stupid newf*g'') and do not date (``normie spotted. real incels are doing this all weekend and have no dates'').
They also are unable to ``ascend'' (i.e., have sex and leave inceldom), and are ``mogged'' by others.
\citet{jaki_online_2019} found similar themes in an earlier incel dataset.

\section{Discussion}
Across our quantitative analysis of the distribution and associations made with identity terms, we see evidence of an ideology where physical appearance determines human value, as has been found with prior work on incels~\citep{maxwell_short_2020,baele_incel_2021,pruden_maintaining_2021}.
This ideology essentializes social constructs, such as race and gender, as biological physical features impacting desirability, with controversy over the role of race.

We find strong evidence for gender as a central focus of incel discussion; mentions of men and women far surpass the number of mentions of any other identity.
We find that this community commonly uses novel identity terms that may not appear in generic lists, including many derogatory terms for women (``foids,'' ``landwhales'').

Increases in mentions of LGBTQ+, political, Jewish, and Black identities, often with stereotypes and conspiracy theories, could suggest this community has incorporated broader far-right trends.
An increasing politicization is reflected in this example post: ``we don't need society to completely accept the incel ideology, we just need to masquerade as normies and keep bashing women, jews and gays.''
Our evidence from text analysis supports the common user movement that \citet{mamie_are_2021} found from manosphere content to alt-right content on YouTube and Reddit. 
We find that many associations on incels.is reinforce stereotypes such as LGBTQ+ identities being fake, Black people being criminals, and antisemitic conspiracy theories.
Users who are central in the forum's shared network devote more identity mentions, proportionally, to Black, LGBTQ+, and Jewish people compared to average users, suggesting that leaders on the platform play a role in broadening the discussion to include mentions of marginalized identity groups other than women.
We also find more mentions of neurodiversity and mental health in this online community than in a dataset of white supremacist online content, which may be part of a victimhood narrative.

The overarching black-pilled ideology of physical appearance determining human worth also extends to talk about incels themselves on incels.is. \citet{jaki_online_2019} also find this ``negative self-image'' on a precursor forum, which is theorized by \citet{nagle_investigation_2015} and \citet{ging_alphas_2019}.
Though incels are presented as occupying the lowest status among men, we find fierce gate-keeping around who can claim the identity and its perceived victimhood.
This echoes theoretical work by \citet{kleinke_intergroup_2015} finding that online communities often disparage less typical members along with out-groups.

Central users are active in discussions of authenticity.
Such victimhood could lead ``authentic'' misogynist incels to pursue symbolic--or material--action against ``fake'' incels in the community but also against the perceived unjust system and the women they believe benefit from it.

Identities constructed in interaction are negotiated~\citep{Bucholtz2010}; we find contention around race in inceldom.
``White'' is associated with both true and fake incels on the platform, often in connection with the folk JBW theory that white men appeal to women of all races.

\paragraph{Implications for automated hate speech detection}
Central to many hate speech definitions is whether a text denigrates groups based on identity characteristics~\citep{sellars_defining_2016,sanguinetti_italian_2018,poletto_resources_2021}.
Identity terms are, thus, a major indicator and concern for hate speech detection.
In our analysis of identity construction on incels.is, we confirm that mentions of men and women identity terms are much more frequent than in a similar source of unlabeled hate speech: white supremacist data.
Incel texts, then, may be a good source of unlabeled or annotated data for misogyny detection.
The dangerous black-pilled ideology in particular is missing from current misogyny datasets~\citep{guest_expert_2021}.
Such data should be considered, but the broader issue is a need for subject matter expertise in building such datasets for automated hate speech detection.
Experts should be consulted to know where to look for training data so that specific types of hateful movements, with lexical or other linguistic innovations, are not overlooked.

We find that almost 30\% of identity mentions in our dataset involve community-specific neologisms, often derogatory terms against women.
Training hate speech classifiers on data that does not include these terms hinders the ability to detect this substantial source of contemporary online misogyny.
% incels and related online male supremacist movements whose lexicons are regularly exported into more mainstream online communities.

Our analysis also draws attention to the ideological associations being made with identities in this discourse space.
We find problematic stereotypes against not only women, but also LGBTQ+, Black, and Jewish people.
Thus, incel text data is not only a source of misogyny, but also reflects broader trends related to the mainstreaming of far-right beliefs.
Particularly pernicious is a black-pilled ideology that physical appearance determines human value, a reinforcement and extension of essentialized gender and racial hierarchies.
Hence, fatphobia, homophobia, ableism, and racism are all wrapped up in misogynist incel content.
Automatically detecting this broader ideology may be unattainable or extremely difficult with machine learning techniques, but we emphasize practitioners and researchers should be aware of this ideology.
A narrow focus on hate against women from these communities will miss these important--and increasing--trends toward politicization and hate against other groups.

\section{Conclusion and Future Work}
The incel movement and the collective identity around it is a relatively new expression of male supremacism.
In this paper, we use quantitative text and network analysis techniques to investigate how identities are constructed in discourse on one of the largest incel forums.
We study the identity group mention frequency over time, as well as actions and attributes associated with them.

We find that talk about women and men dominates identity mentions on this forum, though mentions of marginalized identities commonly targeted by far-right groups increase from 2017-2021, appearing in textual contexts that propagate stereotypes.
Many of these mentions use novel, community-specific identity terms that would be missed with generic lists of identities or hate speech training data from other contexts.
Future work could systematically evaluate the ability of existing hate speech classifiers to handle this jargon, as well as the particularly dangerous black-pilled ideology.

This ideology is apparent in discussions of identities, including in-group ones, that reinforce rigid physical hierarchies based on attractiveness, gender, and race. 
We find race is a site of contention in discussions of who are ``true'' incels.
Gatekeeping around incel authenticity is common.

Negotiation around race and inceldom, as well as intersectional racism and misogyny in incel forums would be a fruitful avenue for future work.
This dataset could also be compared with other incel discussions, such as incels.me and earlier banned subreddits r/incels and r/braincels.
The role of platform affordances and informal mentorship on the platform could be further investigated, as \citet{perry_cognitive_2021} mapped different user pathways in and out of r/TheRedPill.
Further network analysis could reveal how the behaviors we identify, including a rise in mentions of marginalized and political identities, were spread in this community and why.

\section*{Limitations}
Our approaches largely focus on explicit mentions of identity terms.
This does not capture whether the identity term is the target of hate speech, which would require further analysis.
This approach also does not capture attitudes held toward high-profile members of those groups, which play a role in circulating associations with identities (such as personal attacks on women in gaming or the use of ``George Soros'' as shorthand for antisemitic conspiracy theories).
Future work may try to capture and measure these attitudes.

Incels.is is a large, popular forum for black-pilled incel discourse, which has a unique and extreme ideology that we argue is under-represented in current hate speech datasets.
However, our analysis is limited to this forum, and the trends we identify may not apply to more moderate incel discourse (e.g., r/IncelsWithoutHate) or related online male supremacist movements, such as MGTOW and PUAs.
Though these communities are known to have related, but distinct jargon~\citep{farrell_use_2020}, we emphasize that researchers should recognize these lexical innovations in their annotations for hate speech and include a variety of these communities in training datasets for misogyny.

\section*{Ethics Statement}
The Association of Internet Researchers (AoIR) acknowledges internet research is complex, dynamic and often involves many gray areas-- specifically related to what constitutes human subjects, private versus public spaces and data versus persons~\citep{markham_ethical_2012}. For this reason, the AoIR guidance recommends an inductive, ongoing and context-specific approach to ethics throughout the research process. At all stages, this involves being mindful of the vulnerability of the community under study and taking efforts to protect them where appropriate, while balancing their rights with social benefits and the researcher's right to conduct research. 

Following this guidance, we subscribe to a utilitarian philosophy where we focus on doing the greatest good for the greatest number of people. In the case of black-pilled incels, we believe the necessity to better understand this potentially dangerous group outweighs the possible damage to forum members. For this reason, in addition to the AoIR guidance outlined above, we have followed some commonly accepted standards to protect participants and refrain from amplifying misogynist voices.

Data was collected only from publicly available online message boards and no private or identifiable information has been included in this manuscript. We are not publishing user names, though we did observe them in our analysis of central users. We also did not subscribe to any channels or recirculate any content to ensure our work does not contribute to the monetization of the forum or associated accounts. Following the WOAH recommendation, we paraphrase posts to retain key aspects while protecting users' privacy. 

\section*{Acknowledgements}
This work was supported in part by the Collaboratory Against Hate: Research and Action Center at Carnegie Mellon University and the University of Pittsburgh. 
The Center for Informed Democracy and Social Cybersecurity at Carnegie Mellon University also provided support.

\bibliography{custom,references,anthology}
\bibliographystyle{acl_natbib}

\appendix

\section{Additional Tables}
\label{sec:appendix}
\begin{table*}[tb!]
\begin{tabularx}{\textwidth}{llX}
% \begin{tabular}{lll}
\toprule
\textbf{Identity} & & \textbf{Top PMI\textsuperscript{3} terms} \\
\midrule
\multirow[c]{3}{*}{Truecels} & \textit{Attr} & biggest, truest, real, actual, giga, ultimate, legit, confirmed, certified, blackpilled, ugly, absolute, genuine, hope, ethnic, white, old, automatic, fellow, bigger, other \\
 & \textit{Act\textsubscript{S}} & ascend, get, post, know, rise, knows, remain, go, looks, confirmed, relate, rot, understand, cope, use, tried, need, browse, make, suicide, spend, ldar, roped, take \\
 & \textit{Act\textsubscript{O}} & pleasure, confirmed, banning, mog, rejected, bluepilled, help, laid, banned, born, save, doomed, over, seen, calling, see, die, excluded, bullying, dude, mock, mocking \\
\cline{1-3}
\multirow[c]{3}{*}{Fakecels} & \textit{Attr} & fucking, larping, biggest, volcel, inb4, obvious, banned, known, defending, other, massive, fuck, tbh, confirmed, gtfo, potential, normie, likely, one, looking, users \\
 & \textit{Act\textsubscript{S}} & detected, gtfo, confirmed, spotted, get, ascend, post, say, need, banned, try, posting, larping, fuck, smh, come, leave, coming, bragging, go, worry, ruining, invade, piss \\
 & \textit{Act\textsubscript{O}} & ban, calling, banned, gtfo, weed, defending, expose, fucking, larping, spot, call, exposed, defends, exposing, smell, purged, banning, defend, confirmed, found \\
\bottomrule
% \end{tabular}
\end{tabularx}
\caption{Representative actions and attributes associated with truecels and fakecels identity group terms in the incels.is dataset. \textit{Attr} refers to attributes, while \textit{Act\textsubscript{S}} are actions for which the identity group is a subject and \textit{Act\textsubscript{O}} are actions for which the identity group is an object.
} 
\label{tab:tf_actions_attributes}
\end{table*}
% Generated in identity_associations.ipynb
Table \ref{tab:tf_actions_attributes} shows actions and attributes associated with ``trucels'' and ``fakecels,'' common incel variants mentioned in incels.is.

\end{document}